%% file: main.tex
\documentclass{styles/svproc}
\input{header.tex}
\usepackage{layouts}

\begin{document}
\mainmatter              
\title{Adaptive Preload Control of Cable-Driven Parallel Robots for Handling Task}
\titlerunning{Adaptive Preload Control of Cable Robots}  
%
\author{Thomas Reichenbach\textsuperscript{\Letter} \and Johannes Clar \and Andreas Pott
\and Alexander Verl}
\authorrunning{Th. Reichenbach et al.} 
%
\tocauthor{Thomas Reichenbach, Johannes Clar, Andreas Pott, Alexander Verl}
\institute{
	\acf{ISW}\\
	University of Stuttgart\\
	Seidenstr. 36, 70174 Stuttgart, Germany\\
	\email{thomas.reichenbach@isw.uni-stuttgart.de}\\
	\texttt{https://www.isw.uni-stuttgart.de/}
}

\maketitle              

\input{acronyms.tex}

%
%

\input{sections/0_Abstract.tex}

%
%
\input{sections/1_Introduction.tex}

%
%
\input{sections/2_AdvancedKinematics.tex}

%
%
\input{sections/3_NullspaceControl.tex}

%
%

\input{sections/4_ExperimentalInvestigation.tex}

%
%

\input{sections/5_Conclusion.tex}

%
%
\input{sections/9_References.tex}
\end{document}

%% file: header.tex
\usepackage{fontenc}
\usepackage{microtype}
\usepackage{xparse}
\usepackage{etoolbox}
\usepackage{url}

\usepackage[nolist,%
]{acronym}
\usepackage{relsize}
\usepackage{cite}
\usepackage{siunitx}
\usepackage{bm}
\usepackage{blindtext}
\usepackage[numbers]{natbib}
\usepackage{amsmath}
\usepackage{amsfonts}
\usepackage{amssymb}
\usepackage{amstext}
\usepackage{array}
\usepackage{mathtools}
\usepackage{tabularx}
\usepackage{hhline}
\usepackage{booktabs}
\usepackage{graphicx}
\usepackage[misc,%
]{ifsym}
\usepackage[compatibility=false,%
]{caption}
\usepackage{subcaption}
\usepackage{xspace}
\usepackage[%
  colorlinks=false,%
  pdfborder={0 0 0},%
  bookmarks=true,%
]{hyperref}
\usepackage[numbered]{bookmark}
\usepackage[%
  noabbrev,%
  capitalize,%
]{cleveref}
\usepackage{enumitem}

\newcolumntype{L}{>{$}l<{$}}%

\newcommand\vect[1]{\bm{#1}}
\newcommand\matr[1]{\bm{#1}}
\newcommand\transp[1]{\ensuremath{#1^{\mkern-1.5mu\top}}}

\newcommand{\norm}[2][2]{\Vert {#2} \Vert_{\smaller[2]{#1}}}

%% file: acronyms.tex

\begin{acronym}
	\acro{CDPR}[CDPR]{cable-driven parallel robot}
	\acroplural{CDPR}[CDPRs]{cable-driven parallel robots}
	
	\acro{DOF}{degrees of freedom}
	\acro{EE}{end effector}
	\acro{IK}{inverse kinematics}
	\acro{ISW}[\textit{ISW}]{Institute for Control Engineering of Machine Tools and Manufacturing Units}
	\acro{RRPM}{redundantly restrained positioning mechanism}
	\acro{WFW}{wrench-feasible workspace}
	\acro{APC}{adaptive preload control}
	\acro{CF}{Advanced Cloed-Form}
\end{acronym}

%% file: sections/0_Abstract.tex
\begin{abstract}
This paper presents a method for dynamic adjustment of cable preloads based on the actuation redundancy of \acp{CDPR}, which allows increasing or decreasing the platform stiffness depending on task requirements.
This is achieved by computing preload parameters with an extended nullspace formulation of the kinematics.   
The method facilitates the operator's ability to specify a defined preload within the operation space.
The algorithms are implemented in a real-time environment, allowing for the use of optimization in hybrid position-force control.
To validate the effectiveness of this approach, a simulation study is performed, and the obtained results are compared to existing methods.
Furthermore, the method is investigated experimentally and compared with the conventional position-controlled operation of a cable robot.
The results demonstrate the feasibility of adaptively adjusting cable preloads during platform motion and manipulation of additional objects.
\end{abstract}

%% file: sections/1_Introduction.tex
\pdfbookmark[1]{1.~Introduction}{Introduction}
\section{Introduction}
The use of serial manipulators is well established for handling tasks in automation and production environments.
However, for large-scale manipulation the workspace and payload of serial robots are limited.
These disadvantages can be overcome by using \aclp{CDPR} (short: cable robots) that use cables instead of rigid prismatic actuators to control an end-effector.
Due to their flexibility and low weight, cables can be stored compactly on drums so that high maximum actuation length and acceleration can be reached.
Thus, a large geometric workspace can be made possible.
A well-known example for a large-scale cable robot is FAST~\cite{Li.2019e}, which is a spherical radio telescope with a span width of 600~\si{\metre} and a payload of approx. 30~\si{\tonne}.
Other examples are the Robotic Seabed Cleaning Platform (RSCP)~\cite{Gouttefarde.2023} for removing marine litter and high rack warehouse solutions~\cite{Khajepour.2023}.
Efficient operation is essential for the handling tasks of robotic systems.
A high accuracy is required for loading and unloading objects, while the motion between handling operations does not require high accuracies, but low energy consumption is necessary.
For cable robots, increasing the platform stiffness leads to a more precise operation and higher energy consumption, whereas decreasing leads to a low energy consumption due to lower cable force and also lower accuracy~\cite{Nguyen.2014c}.
High energy consumption can be assumed by applying high cable forces on the platform, which require high motor torques and therefore also high motor currents.
Conversely, low energy consumption can be achieved with low cable forces.

Parallel cable robots are redundant mechanisms with number of actuated cables $m \ge n+1$, whereby the platform can be manipulated with $n$ \acp{DOF}. Verhoeven~\cite{Verhoeven.2004} already describes that cable robots have the potential to increase or decrease the cable preload and consequently adjusting the platform stiffness by exploiting the nullspace.
Later, Kraus et al.~\cite{Kraus.2013c} presented an energy efficient computation method of the force distribution of a cable robot.
Therefore, a gradient vector is used to compute a reference force and subsequently solve the static equation of a cable robot with the Moore-Penrose pseudo inverse.
It has been proven that adjusting the cable force results in lower energy consumption than serial kinematics with a similar operation range.
Disturbances due to changes during handling tasks, i.e. loading or unloading an object, and the pose error of the platform were not investigated. 
Furthermore, Fabritius et al.~\cite{Fabritius.2023} presented a correction method for cable forces using nullspace to keep the cable forces within the predefined limits and to compensate disturbances within the cable acutation system.
However, the cable forces and consequently the platform stiffness cannot adaptively change during motion.
Other relevant works are \cite{Guagliumi.2024,Reichenbach.2021,Santos.2020,Kraus.2015} in which hybrid position-force controllers of cable robots are mainly investigated.
However, the aim of the presented control concepts is not to control the cable preload in nullspace and thus adaptively adjust the platform stiffness, but to reduce disturbing effects caused by the cable actuation system or the environment.
Hence, in this paper, an \ac{APC} for redundantly restraint cable robots which allows adjustment of the cable preload during motion and during object manipulation is presented.

%% file: sections/2_AdvancedKinematics.tex
\pdfbookmark[1]{2.~Advanced Kinematic}{Advanced Kinematic}
\section{Advanced Kinematics}
\pdfbookmark[2]{2.1.~Geometric Relations}{Geometric Relations}
\subsection{Geometric Relations}
\begin{figure}
	\centering
	\includegraphics[width=0.6\linewidth]{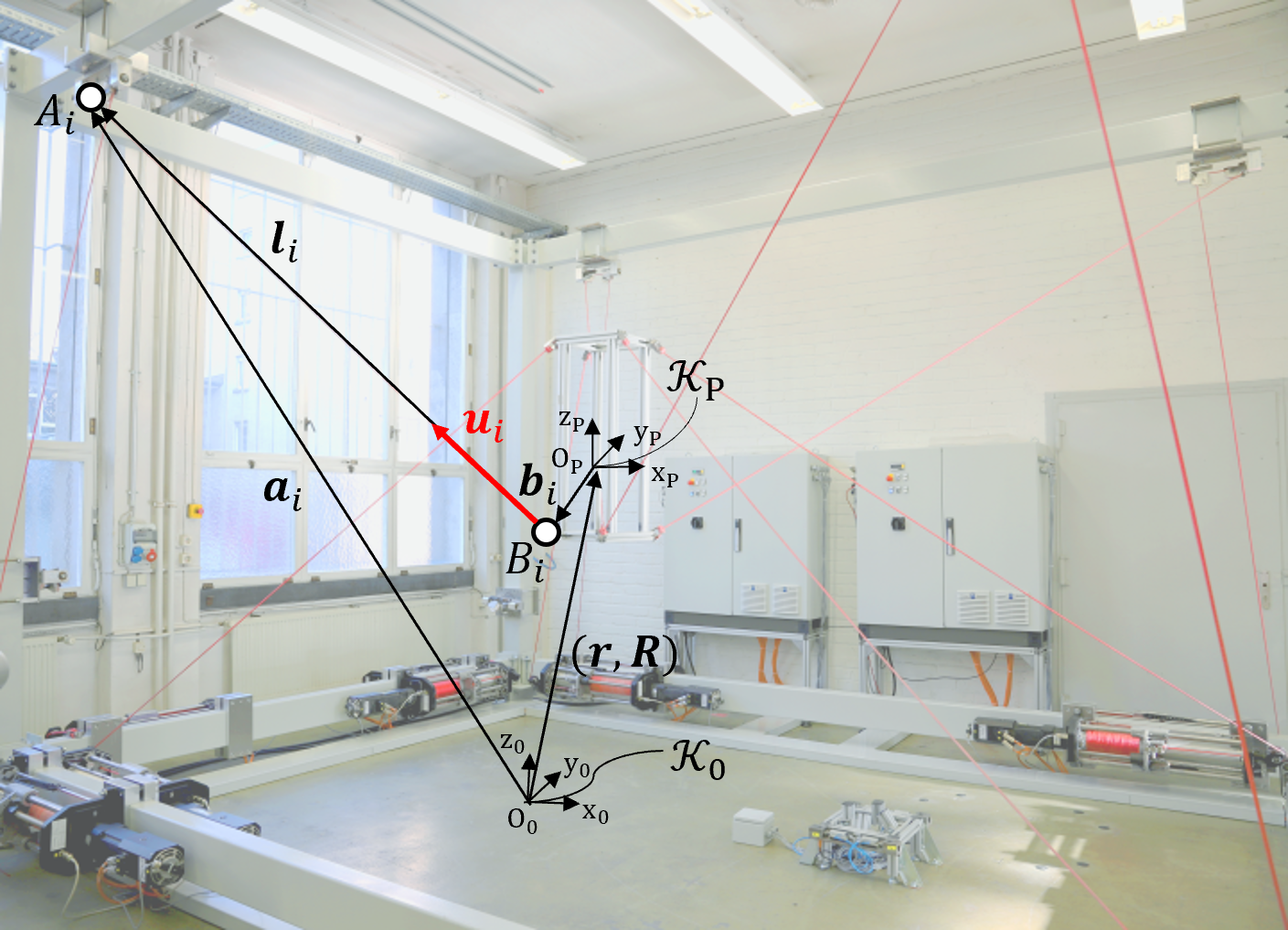}
	\caption{%
		Algebraic loop of inverse kinematics of \textit{COPacabana} robot at \acs{ISW}~\cite{Trautwein.2020}. With $A_i \in \mathcal{K}_0$ as frame anchor points and $B_i\in \mathcal{K}_\text{P}$ as platform anchor points.%
	}
	\label{fig:1}
\end{figure}

In~\cref{fig:1}, the geometric relations of a cable robot are deduced from the algebraic loop of the inverse kinematics with the example of the $i$-th cable. This leads to
\begin{align}
	l_{i}
	& = \norm{\vect{l}_{i}}
	= \norm{ \vect{a}_{i} - \vect{r} - \matr{R} \vect{b}_{i} } \, , \label{eq:1}
\end{align}
where $\vect{l}_{i}$ is the vector of the $i$-th actuated cable with $l_{i}$ as its length, $\vect{a}_{i}$ the frame anchor points, $\vect{r}$ the platform position, $\matr{R}$ the platform rotation matrix, and $\vect{b}_{i}$ the platform anchor points within the local platform coordinate system.

The so-called structure matrix $\transp{\matr{A}}$ is computed by normalizing inverse kinematics for each cable $i$ with given platform pose $(\vect{r}, \matr{R}) \in \text{SE}(3)$ with the Euclidean norm $\norm{ \cdot}$.
The structure matrix can also described as the negative transposed Jacobian of direct kinematics $\transp{\matr{A}} = -\transp{\matr{J}}_{\text{DK}}$ which has the cable direction $\vect{u}_{i} = \frac{ \vect{l}_{i} }{ \norm{ \vect{l}_{i} } }$ and twist with respect to a reference coordinate system $\matr{R} \, \vect{b}_{i} \times \vect{u}_{i}$ as matrix entries,
\begin{align}
	\transp{\matr{A}}
	& = -\transp{\matr{J}}_{\text{DK}} =
	\begin{bmatrix}
		\vect{u}_{1}
		& \cdots
		& \vect{u}_{m} \\
		\matr{R} \, \vect{b}_{1} \times \vect{u}_{1}
		& \cdots
		& \matr{R} \, \vect{b}_{m} \times \vect{u}_{m}
	\end{bmatrix} \, . \label{eq:2}
\end{align}
For the technical use of running cables, the use of deflection pulleys are common, so that the inverse kinematics can be extended with pulley kinematics, as described in \cite{Pott.2012}.
This extended pulley kinematics is used for the simulative and experimental investigations in this paper.

\pdfbookmark[2]{2.2~Nullspace Extension}{Nullspace Extension}
\subsection{Nullspace Extension}
The dynamics of the mobile platform is summarized to wrench $\vect{w}_\text{c}$ such that $\transp{\matr{A}}\vect{f}=\vect{w}_\text{c}$ is fulfilled, where $\vect{f} = [f_1,\ldots,f_{m}]$ are the applied cable forces due to joint-space actuation.
Because the cable robot platform is actuated with at least $m \ge n+1$ cables, the structure matrix $\transp{\matr{A}}$ is non-quadratic and not invertible.
Thus, the kernel of the structure matrix $\mathrm{ker}(\transp{\matr{A}}) = \{\vect{v} \in \mathbb{R}^m \mid \transp{\matr{A}} \vect{v}_j = 0 \}$ is not empty, assuming that the robot is not in a singular pose.
For example, a redundancy of $\rho = m-n$ leads to a pose dependent nullspace with size $(m \times \rho)$ such that $\matr{N} = [\vect{v}_1, \ldots,\vect{v}_\rho]$.
The main approach of the \ac{APC} which is presented in this paper is adjusting the geometric stiffness of the platform on the platform and identify a set of preload parameters such that a solution of force distribution is found.
To do so, cable force are measured on the frame (i.e. with pulleys) and converted to the applied wrench $\vect{w}_\text{c}$. 
The stiffness of a cable robot is represented by two parts: the pose-dependent stiffness applied due to cable elongation and the geometric stiffness which depends on the applied cable forces and pose (see Kraus~\cite{Kraus.2015}).
To adjust the geometric stiffness, the non-quadratic structure matrix $\transp{\matr{A}}$ will be extended by its transposed nullspace, such that
\begin{align}
	\transp{\matr{A}_\text{adv}} 
	& = 
	\begin{bmatrix}
		\transp{\matr{A}}\\
		\transp{\matr{N}}
	\end{bmatrix}
	. \label{eq:4}
\end{align}
Hence, $\transp{\matr{A}_\text{adv}}$ is an invertible quadratic matrix which can be used to compute a unique force distribution for a given platform wrench with
\begin{align}
	(\transp{\matr{A}_\text{adv}})^{-1} \vect{w}_\text{c,adv} =  \vect{f} \, 
	, \label{eq:5}
\end{align}
where $\vect{w}_\text{c,adv}=\transp{[\vect{w}_\text{c},\vect{\lambda}]}$, with preload parameters $\vect{\lambda}=\transp{[\lambda_1,\ldots,\lambda_\rho]}$.

\pdfbookmark[2]{2.3~Preload Parameter Optimization}{Preload Parameter Optimization}
\subsection{Preload Parameter Optimization}
Depending on the platform pose $(\vect{r},\matr{R})$ and the orientation of the nullspace, feasible preload parameters $\vect{\lambda}$ must be determined.
Furthermore, the cable force should neither drop below a specified minimum nor exceed a maximum limit.
The cable force limitations result from the mechanical properties of the cable robot.
For example, depending on the size of the robot and cable type a specific minimum force $\vect{\mathrm{f}}_\text{min}$ must be applied to the cable to avoid uncontrolled motion of the platform due to lag of stiffness.
The maximum cable force $\vect{\mathrm{f}}_\text{max}$ should not be exceeded because of cable damage or overload of the cable actuation system.
For handling tasks, the platform is moved with or without objects, whereby high stiffness is required during the loading or unloading process.
The movement along the planned trajectory between the two operations must be efficient, whereas a high stiffness is not necessary.
The geometrical stiffness can be seen as a measure of the resistance of the platform, which influences the positioning accuracy~\cite{Nguyen.2014c}.
Hence, it is necessary to adaptively change the preload of the platform with the force distribution of the cables.
Therefore, the preload control parameter $\eta_\text{c}$ is introduced, such that the platform is specifically preloaded for each pose of a tajectory.
Consequently, these conditions can be formulated as the following minimization problem
\begin{alignat}{3}
	\min_{\vect{\lambda}} 
	&
	\quad
	&&
	e =\norm{\underbrace{\matr{N}\vect{\lambda}}_{\vect{f}_\text{tar}} - (\eta_\text{c}\vect{\mathrm{f}}_{\text{max}}+(1-\eta_\text{c})\vect{\mathrm{f}}_{\text{min}})} \label{eq:6}\\
	\text{subject to: }
	&
	\quad
	&&
	\underbrace{(\transp{\matr{A}_\text{adv}})^{-1} \vect{w}_\text{c,adv}}_{\vect{f}_\text{tar}}-\vect{\mathrm{f}}_{\text{max}} \leq 0 \label{eq:7}\\
	&
	\quad
	&&
	\vect{\mathrm{f}}_{\text{min}}-\underbrace{(\transp{\matr{A}_\text{adv}})^{-1} \vect{w}_\text{c,adv}}_{\vect{f}_\text{tar}} \leq 0	 \label{eq:8}\\
	&
	\quad
	&&
	0 < \eta_\text{c} < 1 \quad , \label{eq:9}
\end{alignat}
where $\vect{f}_\text{tar}$ describes the target force distribution for the force controller.
The preload control parameter $\eta_\text{c}$ can be adapted between 0\si{\percent} and 100\si{\percent} during motion.
Thus, a real-time capable optimization algorithm with nonlinear constraints is used to compute feasible preload parameters $\vect{\lambda}$.

%% file: sections/3_NullspaceControl.tex
\pdfbookmark[1]{3.~Adaptive Preload Controller}{Adaptive Preload Controller}
\section{Adaptive Preload Controller}
\begin{figure}
	\centering
	\includegraphics[width=0.9\linewidth]{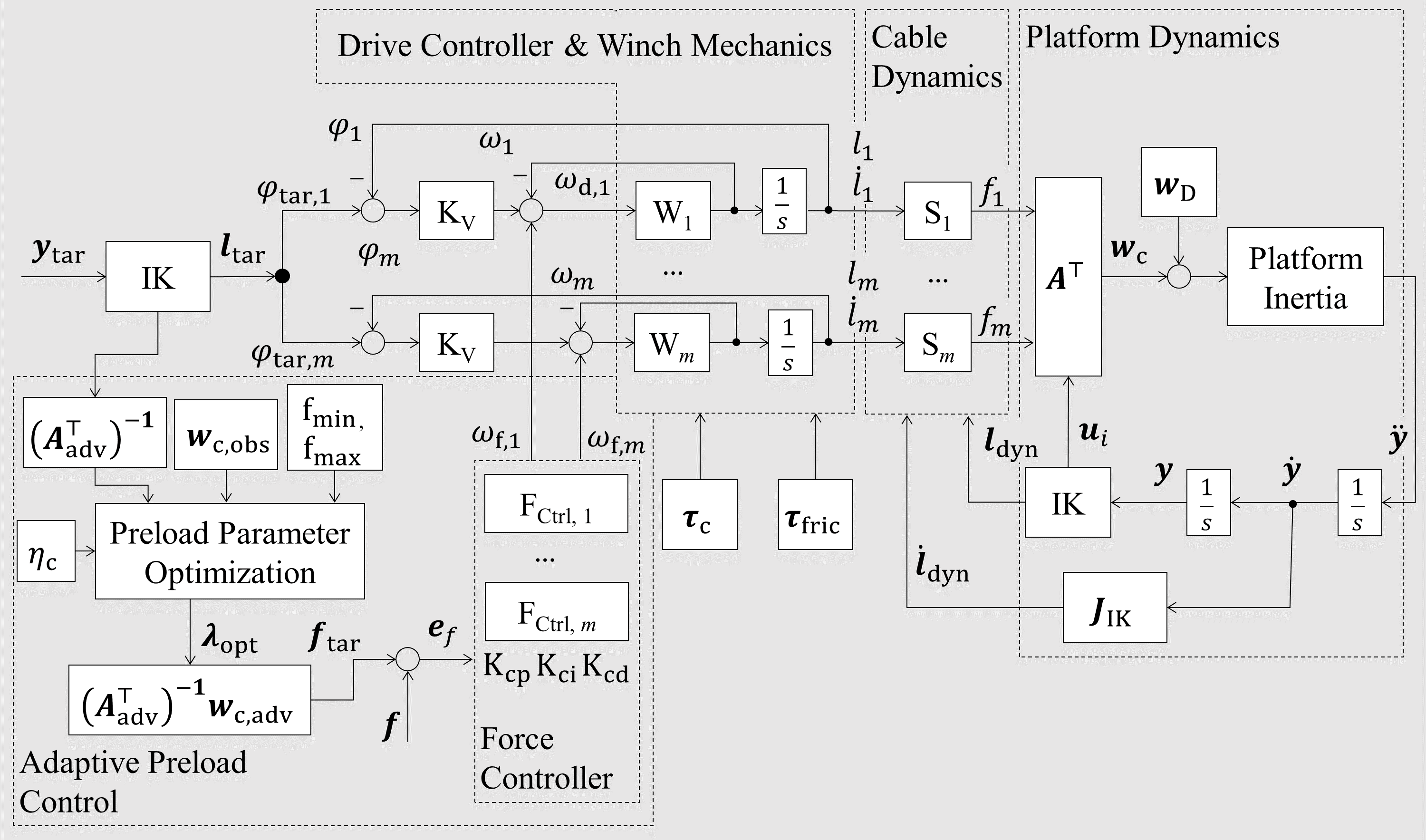}
	\caption{%
		Schematic overview of the implementation of \ac{APC} within the cable robot system (adapted from~\cite{Reichenbach.2021}).%
	}
	\label{fig:2}
\end{figure}

%
To give an overview, \cref{fig:2} shows a schematic block diagram of the \ac{APC} and the relation between the platform and drive systems of the cable robot.
The basic structure of the control system comes from Reichenbach et al.~\cite{Reichenbach.2021}, and is now adapted for the integration of the APC.
The above-mentioned work also includes a detailed description of the modeling of drive controller, winch mechanics, cable dynamics, and platform dynamics.
Hence, the following section focus on \ac{APC}.

To achieve linear actuation using a cable, the target pose $\vect{y}_{\text{tar}}=\transp{[x,y,z,\alpha,\beta,\gamma]}$ of the platform, which includes Cartesian position $\vect{r}_\text{tar}$ and transformed orientation $\vect{R}_\text{tar}$ specified in Euler angles, requires the calculation and transformation of the target cable length $l_{\text{tar},i}$ into angular positions $\varphi_{\text{tar},i}$ for the drive systems.
To do so, the \acf{IK}, as defined in~\cref{eq:1}, is used with pulley kinematics~\cite{Pott.2012}, taking into account the transmission ratio of the planetary gear $\textrm{t}_\text{G}$ and the winch $\textrm{t}_\text{W}=2\pi \mathrm{R}_\text{D}$ with $\mathrm{R}_\text{D}$ as nominal drum radius.
Thus, the position controller output is the angular velocity target value
\begin{equation}
	\omega_{\mathrm{q},i}  =  \textrm{K}_\textrm{v} \textrm{t}_\textrm{W} \textrm{t}_\textrm{G} (l_{\mathrm{d},i}-l_i) = \textrm{K}_\textrm{v}(\varphi_{\mathrm{d},i}-\varphi_i),%
	\label{eq:10}%
\end{equation}
with $\varphi_i$ for the actual motor position, $\textrm{K}_\textrm{v}$ as position controller gain.
For preload parameter optimization, the actual wrench on the platform is observed by transforming the actual cable forces with the structure matrix, such that $\vect{w}_\text{c,obs}=\transp{\matr{A}}\vect{f}$.
Consequently, friction forces, i.e. within the cable guidance system, are also transformed into the operational space.
The inverted advanced structure matrix~$(\transp{\matr{A}_\text{adv}})^{-1}$ is computed with the target pose $\vect{y}_\text{tar}$ according to~\cref{eq:2,eq:4}.
Both, the actual wrench and the inverted advance strucutre matrix are computed in real-time for each target pose along a planned trajectory.
Additionally, the adaptive preload control parameter $\eta_\text{c}$ and cable force limits $\mathrm{f}_{\text{lim}}=\transp{[\mathrm{f}_{\text{min}}, \mathrm{f}_{\text{max}}]}$ are specified.
To preload the cables within the cable robots nullspace, a force controller is implemented.
Using the optimized preload parameters~$\vect{\lambda}_{\text{opt}}$ and the advanced structure matrix~$\transp{\matr{A}_\text{adv}}$, target cable forces can be computed with~\cref{eq:5} and controlled with a PID-controller $\textrm{F}_{\text{Ctrl},i}$ for each motor.
The output of the force controller is an additive angular speed value $\omega_{\mathrm{f},i}$.
Thus, the angular speed target for the drive control system consists of the position controller part and the force controller part which leads to $\omega_{\text{tar},i} = \omega_{\mathrm{q},i} + \omega_{\mathrm{f},i}$.

\pdfbookmark[2]{Simulation Study}{Simulation Study}
\subsubsection{Simulation Study}\label{sec:3.2}
To show that the algorithms of the \ac{APC} are correctly implemented, a simulation study is performed.
Therefore, the cable force computation of the presented \ac{APC} is compared with the \ac{CF} method~\cite{Pott.2009}.
The simulation study was performed with the geometrical data of the cable robot \textit{COPacabana} which is presented by Trautwein et al.~\cite{Trautwein.2020}.
However, modifications were made to the frame and platform geometry due to reconstruction.
Thus, the actual data of \textit{COPacabana} is published in~\cite{Reichenbach.2024} as XML file including geometry, drive and control parameters of the cable robot.
For the computation of the force distributions, the Python integration WiPy of WireX is used which is an open-source software environment to develop cable robots and presented by Pott~\cite{Pott.2019}.
To solve~\cref{eq:6}, optimization algorithms COBYLA and SLSQP are investigated. COBYLA uses local derivative-free optimization and SLSQP uses local gradient-based optimization.
Both algorithms produce similar results, and only minor changes can be observed due to numerical noise and different treatment of the termination criteria.
The feasibility of the algorithms is calculated for the force distribution at poses along a path with $r_\mathrm{z} = [0, 3]~\si{\meter}$ and a discretization of 1~\si{\milli\meter}, which is an upward movement of the platform without rotation. 
Thereby, the algorithms are verified whether the solutions of the computed target forces show discontinuities and whether the force distribution of the respective pose is valid.
\begin{figure}
	\centering
	\includegraphics[width=0.9\linewidth]{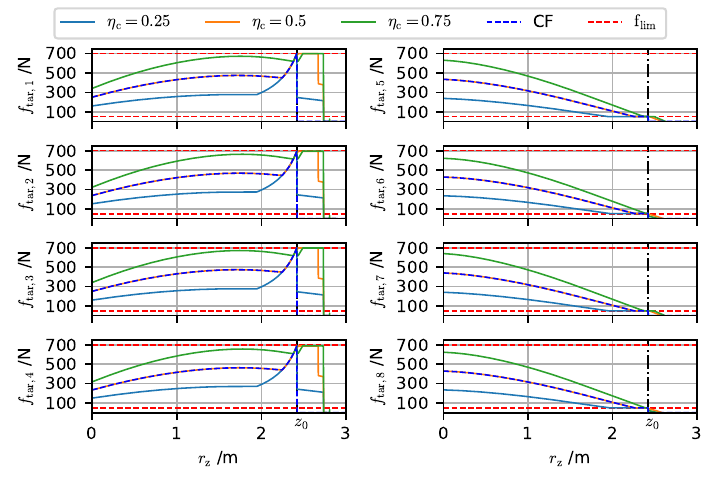}
	\caption{%
		Computed cable force distributions of APC compared to \ac{CF} method with a motion along $r_\mathrm{z} = [0, 3]~\si{\metre}$ and cable force limits $\mathrm{f}_{\text{lim}}=\transp{[50, 700]}\si{\newton}$.
	}
	\label{fig:3}
\end{figure}

The results of the simulation study in~\cref{fig:3} show, that the approach using a preload control parameter $\eta_\text{c}$ to increase or decrease the geometric stiffness of the platform is feasible.
It can be seen that despite different parameters, the optimization algorithm converges until the pose $z_0 = 2.4~\si{\metre}$ is reached, marked with a black line in the figure.
In addition, the results show that a preload control parameter of $\eta_\text{c}=50\%$ correspond to the results of the \ac{CF} method.

%% file: sections/4_ExperimentalInvestigation.tex
\pdfbookmark[1]{4.~Experimental Investigation}{Experimental Investigation}
\section{Experimental Investigation}
\begin{figure}
	\centering
	\includegraphics[width=0.95\linewidth]{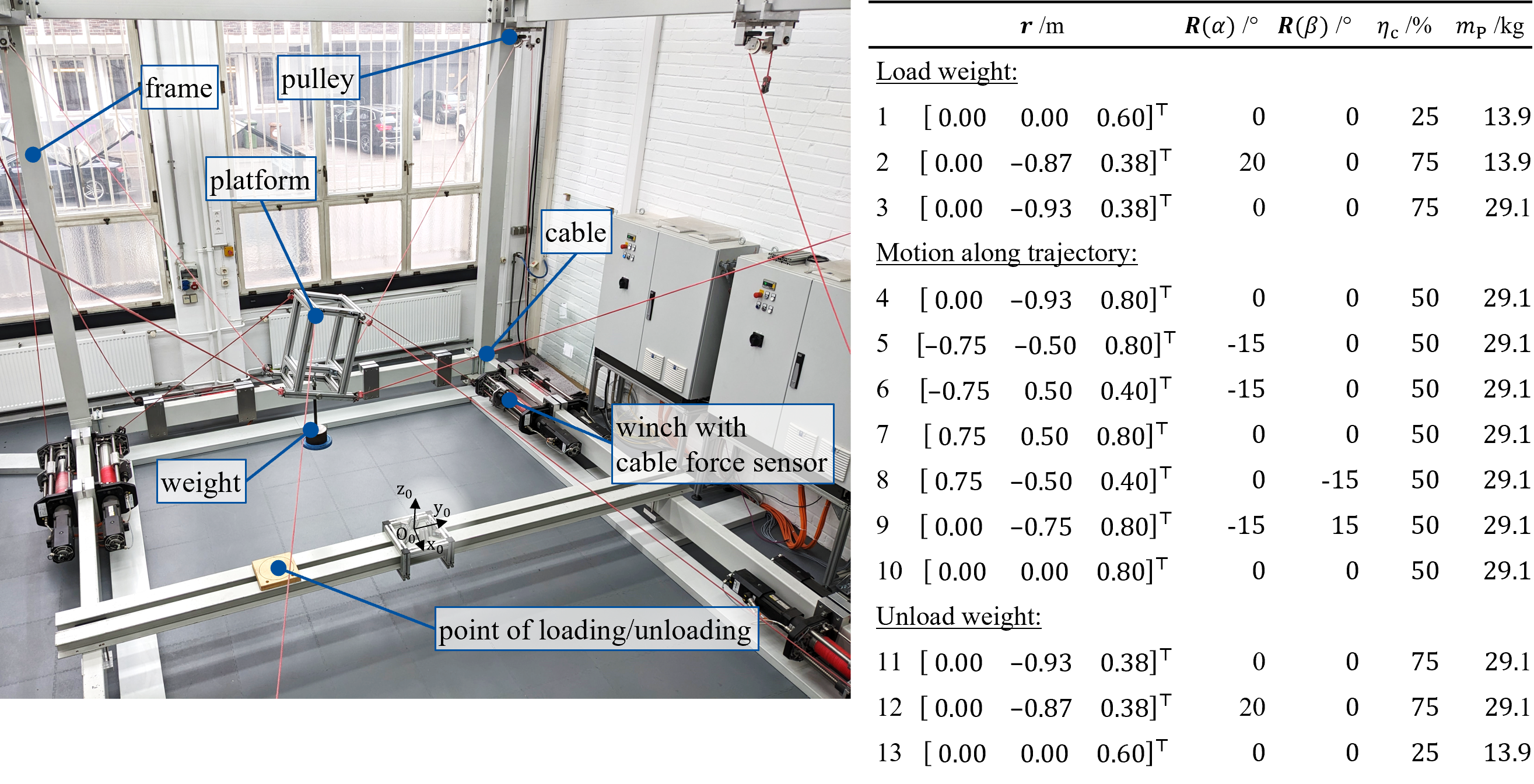}
	\caption{%
		Experimental setup on the left and schematic representation of the performed motion on the right.%
	}
	\label{fig:4}
\end{figure}

%
The experimental investigation is performed on the cable robot \textit{COPacabana} as described in~\cref{sec:3.2} and thus the same parameters are used.
To compute cable force target values for the drive control system, the SLSQP algorithm is converted into real-time capable driver object (TcCOM) in Beckhoff TwinCAT~3 (3.1.4024.54) environment.
This setup is able to compute the nullspace, the optimization, and the force control algorithms within 20~\si{\milli\second} (with a single core of an Intel Core i7-7700 processor).  
In order to validate the \ac{APC} for a handling task experimentally, a weight with $m_{\text{Obj}}=15.2~\si{\kilo\gram}$ is loaded via a load hook onto the cable robot, moved along a trajectory with different rotations and then unloaded again.
The overall weight is $m_{\text{P}}=m_{\text{EE}} + m_{\text{Obj}}$ with platform mass $m_{\text{EE}}=13.9~\si{\kilo\gram}$.
The procedure is performed with conventional position controlled operation using \ac{IK} (STD) and repeated with activated \ac{APC}.
The experimental setup and the platform trajectory are shown in~\cref{fig:4} and video recordings of the two experiments are also published in~\cite{Reichenbach.2024}.
The trajectory during manipulation was performed with a target path velocity of $\dot{\vect{y}}_\text{tar} = 0.34~\si{\metre\per\second}$.

\pdfbookmark[2]{Results and Discussion}{Results and Discussion}
\subsubsection{Results and Discussion}
\begin{figure}[h!]
	\centering
	\includegraphics[width=0.95\linewidth]{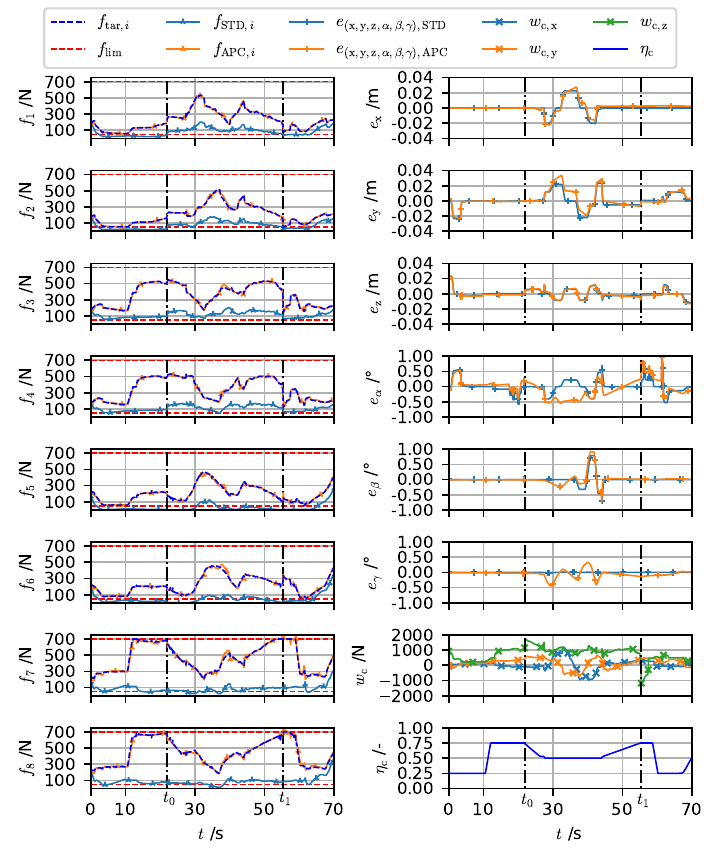}
	\caption{%
		Experimental results with cable forces on the left and pose errors on the right. In addition, measured translational wrenches and preload control parameters during motion are shown.
	}
	\label{fig:5}
\end{figure}

The results of the experiments are shown in~\cref{fig:5}.
On the left, the cable forces of the experiments are shown, where the cable force target values $f_{\text{tar},i}$ are computed with~\cref{eq:5}.
The cable force limits are $\mathrm{f}_{\text{lim}}=\transp{[50, 700]}\si{\newton}$.
The measured cable forces during position controlled operation are $f_{\text{STD},i}$, whereas cable forces of \ac{APC} operation are $f_{\text{APC},i}$.
On the right, position and rotation errors of the platform are shown to validate that changes in geometric stiffness do not lead to large changes in position and rotation, and stay within the tolerance STD operation.
The platform pose~$\vect{y}_{\text{DK}}$ was determined by solving forward kinematics using measured motor encoder position~$\vect{\varphi}$.
Thus, pose errors due to cable elongation between frame and platform anchors cannot be investigated.
The pose errors $\vect{e}=\vect{y}_{\text{tar}}-{\vect{y}_{\text{DK}}(\vect{\varphi})}$ are computed for platform position and rotation $\vect{e}=\transp{[e_\mathrm{x}, e_\mathrm{y}, e_\mathrm{z}, e_\mathrm{\alpha}, e_\mathrm{\beta}, e_\mathrm{\gamma}]}$ for both experiments, STD and \ac{APC} operation.
At $t_0=22~\si{\second}$ the weight is loaded by the platform and at $t_1=55.4~\si{\second}$ the weight is unloaded.

For \ac{APC}, the measured cable forces follow the target cable forces continuously and stay within force limits,  even if the preload control parameter $\eta_\text{c}$ is changed during motion.
While unloading the weight at $t_1$ the cable forces of the lower cables $i=[5,6,7,8]$ exceed the maximum cable force limit for  about 0.5~\si{\second}, as the weight was jammed when it was unloaded. 
Furthermore, minor control errors of the cable force occur during the movement between $t_0$ and $t_1$, which can be explained by the incomplete identification of the wrench $\vect{w}_\text{c}$ from the force sensors on winches and the swinging weight.
This effect can also be observed in the minor increased rotation error $e_\mathrm{\gamma}$.
Comparing the cable forces of STD and \ac{APC} operation, the cable forces during STD operation often fall below the minimum force, which leads to a loss of platform stiffness and reduced position accuracy.
Additionally, there only small differences found between STD and \ac{APC} operation of the computed pose errors.
Hence, the geometrical stiffness of the platform can be adaptively adjusted without loss of pose accuracy within the tolerance of STD operation.
The point in time of loading and unloading the weight can be observed by identified wrench $w_\mathrm{c,z}$.
The swinging effect of the weight during the trajectory can also be observed with the wrenches $w_\mathrm{c,(x,y)}$.

%% file: sections/5_Conclusion.tex
\pdfbookmark[1]{5.~Conclusion}{Conclusion}
\section{Conclusion}
In this work the redundancy of \acsp{CDPR} is exploited to adaptively increase or decrease the geometric stiffness of the platform.
The presented method allows an operator to increase the stiffness of the platform if required when loading and unloading objects.
In contrast, the stiffness of the platform and the energy consumption for valid force distribution can be reduced by the operator, respectively.
Therefore, the developed concept of the \ac{APC} is investigated in a simulation study.
The results are compared to the state-of-the-art \acl{CF} method for cable force distribution computation.
Subsequently, the concept of \ac{APC} is validated within experimental investigation and compared to conventional position controlled operation of a cable robot.
Finally, it is shown that the preload of the cables can be adaptively changed during platform motion and during manipulation of an additional object without changes of the platform pose.
In the future, the accuracy of the cable robot will be investigated in detail with an absolute coordinate measuring system (laser tracker) to show the advantages of \ac{APC} with hybrid position-force control of cable robots.
Additionally, friction disturbances within the drive system will be further investigated in order to save costs on additional cable force sensors.
Furthermore, the developed algorithms can be used for the design and control of highly redundant reconfigurable cable robots.
\vspace{1mm}

\pdfbookmark[1]{Acknowledgments}{Acknowledgments}
{\smaller[1]
\noindent \textbf{Acknowledgments.} This work was supported by the German Research Foundation (DFG-project numbers: 470674716 and 317440765) at the University of Stuttgart.}

%% file: sections/9_References.tex

\pdfbookmark[1]{References}{References}
\renewcommand{\bibsection}{\section*{References}}
{\smaller[2]
\bibliographystyle{styles/bibtex/spbasic}
\bibliography{references}}